\documentclass[10pt,twocolumn,letterpaper]{article}

\usepackage[final]{cvpr}      
\usepackage{graphicx}
\usepackage{booktabs}
\usepackage{multirow}
\usepackage{bbding}
\usepackage{float}

\definecolor{cvprblue}{rgb}{0.21,0.49,0.74}
\usepackage[pagebackref,breaklinks,colorlinks,allcolors=cvprblue]{hyperref}

\def\paperID{4869} 
\def\confName{CVPR}
\def\confYear{2025}

\title{Frequency-domain Learning with Kernel Prior for Blind Image Deblurring}

\author{Jixiang Sun$^1$\footnotemark[1] \quad\quad Fei Lei$^2$\footnotemark[1]\quad\quad Jiawei Zhang$^2$\quad\quad Wenxiu Sun$^2$\quad\quad Yujiu Yang$^1$\footnotemark[2]\\
$^1$Tsinghua University \quad\quad $^2$Sensetime Research\\
}

\begin{document}
\maketitle
\renewcommand{\thefootnote}{\fnsymbol{footnote}}
\footnotetext[1]{Equal contribution.}
\footnotetext[2]{Corresponding author.}
\renewcommand{\thefootnote}{\arabic{footnote}}
\begin{abstract}
While achieving excellent results on various datasets, many deep learning methods for image deblurring suffer from limited generalization capabilities with out-of-domain data. This limitation is likely caused by their dependence on certain domain-specific datasets. To address this challenge, we argue that it is necessary to introduce the kernel prior into deep learning methods, as the kernel prior remains independent of the image context. For effective fusion of kernel prior information, we adopt a rational implementation method inspired by traditional deblurring algorithms that perform deconvolution in the frequency domain. We propose a module called Frequency Integration Module (FIM) for fusing the kernel prior and combine it with a frequency-based deblurring Transfomer network. Experimental results demonstrate that our method outperforms state-of-the-art methods on multiple blind image deblurring tasks, showcasing robust generalization abilities. Source code will be available soon.
\end{abstract}
\section{Introduction}
\label{sec:intro}
Image deblurring stands as a classic and important problem in the realm of low-level vision. Image deblurring aims to recover a high-quality sharp image from its blurry counterpart. Due to its inherently ill-posed nature, image deblurring has always been presented as a challenging task. The problem format of this task is roughly constructed as $\mathbf{y} = \rm Blur(\mathbf{x}, \mathbf{k}) + \mathbf{n}$, where Blur(,) indicates a blurring operation, conventionally modeled as a convolution operation. $\mathbf{x}$ denotes the original sharp image, $\mathbf{k}$ represents the convolution kernel responsible for inducing blurriness, and noise $\mathbf{n}$ is typically attributed to the inherent imperfections of the camera lens, introducing further complexity to the deblurring task. 

In light of the development of deep neural networks, multiple approaches to image deblurring have been proposed in recent years. Recent methods include convolutional neural networks (CNNs)~\cite{schuler2015learning, sun2015learning, chakrabarti2016neural, kaufman2020deblurring}, and attention-based solutions~\cite{shen2019human, purohit2020region}.
Despite that they have achieved considerable results, these methods usually fail to generate satisfying results when meeting out-of-domain data. Specifically, they suffer from severe performance drop with pretrained weights on other datasets. In contrast, some traditional deblurring methods~\cite{krishnan2009fast, kruse2017learning, ren2017video} solve the problem in a closed-form way, suitable for any image content and dataset distribution. This is partly because non-deep learning methods are able to utilize the information of blur kernel for image deblurring, which is independent of image content. In this case, we suggest that bringing blur kernel prior information into deblurring networks may help with its deblurring abilities, especially on out-of-domain data.

The first step is to make an estimation of the blur kernel in advance. Early approaches to solving this problem usually seek to estimate a uniform blur kernel ~\cite{schuler2015learning, ren2020neural}, based on the assumption that motion blur is spatially invariant. These methods show their shortcomings when encountered with real-world blurry images, which usually consist of multiple objects and complex structures, greatly enlarging the variety of motion in a single image. Subsequent works seek to estimate non-uniform blur kernels, for the purpose of processing real-world images.

A number of recent works~\cite{carbajal2023blind, li2022learning} have combined kernel features with deep neural networks, achieving remarkable performance on both synthetic and real-world deblurring tasks. The methods mentioned above have leveraged kernel information in a variety of ways, including direct deconvolution, and learning an adaptive module based on direct channel multiplication and concatenation. However, these incorporation approaches are either restrained to ideal assumptions of image composition or suffering from its rigidness. 

Motivated by the inherent property of Fourier Transform and deconvolution theorem, we propose a novel way of incorporating blur kernel prior, called \textbf{F}requency \textbf{I}ntegration \textbf{M}odule (FIM). Given two functions, one can obtain the result of their convolution by calculating the Hadamard product of their corresponding Fourier transforms, this is because convolution in the spatial domain equals point-wise multiplication in the frequency domain. Since a blurry image is obtained by convolution operation between a sharp image and its matching blur kernels, the blur process can be expressed as multiplication in the frequency domain. Based on the assumption, we propose the kernel-assisted deblurring module FIM. In order to optimize the integration of the proposed module with the network, we adopt an end-to-end Transformer-based deblurring network architecture based on frequency domain as the backbone of our work~\cite{kong2023efficient}. Due to the fact that the backbone has an inherent asymmetric structure, we explore different structures of our module, along with different ways of implementing it. Ablation experimental results show that with a residual connection inside the block, and both encoder and decoder insertion, the proposed FIM block can best enhance the network's abilities in image deblurring.

Compared with traditional end-to-end deblurring networks, the innovation of blur kernel prior brings in extra crucial information about the degradation of an image. An advantage of blur kernels is that they are irrelevant to the content of the image. In this way, the learned network will not only be able to reconstruct sharp images based on domain-specific image information, but will also take into account the domain-unrelated blur kernel information. Therefore, we argue that kernel-based deblurring networks may achieve better performance on out-of-domain data. Experimental results prove our assumption, demonstrating that our proposed method helps with enhancing the blur removal process, achieving state-of-the-art on multiple image deblurring missions, both on in-domain data and out-of-domain data.

Our main contribution can be summarized as follows:
\begin{itemize}
\item We develop a novel way of utilizing blur kernel prior by fusing frequency domain information to enhance image deblurring, and conduct experiments to show the effect of kernel prior on image deblurring tasks.
\item We tackle the problem of out-of-domain data distribution by introducing an explicit blur kernel prior. Experiments on out-of-domain benchmark datasets show the effectiveness of our proposed module.
\item We explore different structures and implementation strategies for our network, and also perform experiments on a non-blind version of our method. Extensive experimental results on benchmark datasets demonstrate that our proposed multi-scale implementation method is capable of improving the performance of deblurring networks, outperforming state-of-the-art methods on this task.
\end{itemize}
\section{Related Work}
\label{sec:related}


\subsection{Kernel-based Deblurring}

Since a blurry image can be obtained by a sharp image with blur kernels and a blurring operation, it is common to utilize the kernel information to assist the deblurring process. The information on the kernel is either known or unknown, depending on the task. Early kernel-based deblurring methods are usually based on the assumption that the information of blur kernels is given. This kind of task is known as non-blind deblurring. Some non-deep-learning methods use natural image priors, such as global image prior~\cite{krishnan2009fast} and local image prior~\cite{zoran2011learning}. To solve the problem of undesired ringing artifacts in traditional methods, Xu et al.~\cite{xu2014deep} and Ren et al.~\cite{ren2018deep} combine spatial deconvolution with deep neural networks. 

With the development of photography, the need for blind image deblurring has greatly increased. A number of blind deblurring methods have been proposed. Ren et al.~\cite{ren2020neural} and Schuler et al.~\cite{schuler2015learning} make the assumption that a blurry image is obtained from its sharp counterpart and a uniform blur kernel. Recent methods pay more attention to estimating non-uniform blur kernels~\cite{sun2015learning, xu2017motion, kaufman2020deblurring, chen2018reblur2deblur, li2022learning, carbajal2023blind}, mostly employing the structure of CNN for the estimation operation. Kaufman et al.~\cite{kaufman2020deblurring} first recover the blur kernel using a dedicated analysis network, they are then fed into a second synthesis network that performs the non-blind deblurring along with the input image and recovers the sharp image. Chen et al.~\cite{chen2018reblur2deblur} fine-tune existing deblurring neural networks in a self-supervised fashion, the method computes a per-pixel blur kernel which is used to reblur the estimated sharp frames back onto the input images. Li et al.~\cite{li2022learning} propose a framework to learn spatially adaptive degradation representations of blurry images along with a joint image reblurring and deblurring learning process. Carbajal et al.~\cite{carbajal2023blind} train a network to estimate both the basic composition of blur kernels and their mixing coefficients, with a deblurring network based on non-blind deconvolution trained jointly. 

\begin{figure*}[ht!]
  \centering
  \includegraphics[width=1.0\linewidth]{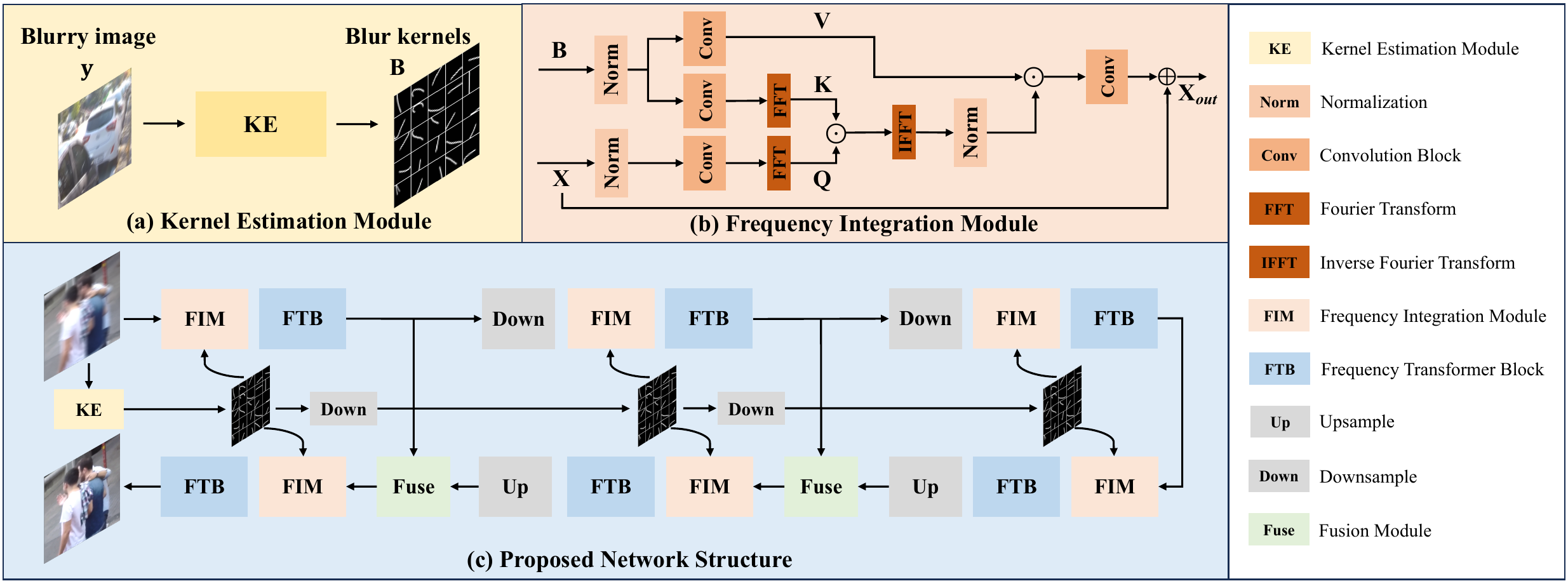}
  \hfill

  \caption{Overview of the network architecture of our proposed method. (a) The structure of the Kernel Estimation Module. (b) The structure of our proposed Frequency Integration Module (FIM). (c) The structure of the full pipeline in our proposed method.}
  \label{fig:overall}
\end{figure*}

\subsection{Frequency-domain based Deblurring}

In contrast to most deblurring methods which focus on spatial domain, in recent years, there have emerged a number of methods that explore the function of frequency domain~\cite{kruse2017learning, mao2023intriguing, cui2023dual, kong2023efficient}. The Fast Fourier transform (FFT) is performed on image pixels or features to extract frequency information. Kruse et al.~\cite{kruse2017learning} propose a simple yet effective boundary adjustment method that alleviates the problematic circular convolution assumption, which is necessary for FFT-based deconvolution. Mao et al.~\cite{mao2023intriguing} extract faithful information about the blur pattern by applying ReLU operation on the frequency domain of a blur image followed by inverse Fourier transform. Cui et al.~\cite{cui2023dual} develop a dual-domain attention mechanism on spatial and frequency domains simultaneously and implement self-attention in the style of dynamic group convolution to reduce computation costs. Kong et al.~\cite{kong2023efficient} replace the matrix multiplication in the attention operation with frequency-domain element-wise multiplication, namely frequency domain-based self-attention solver (FSAS), and develop a frequency domain-based feed-forward network for better image deblurring, not only reducing the complexity of the network, but also achieving state-of-the-art on this task.

\section{Method}
\label{sec:method}

 We propose a framework with (i) a kernel estimation module that estimates spatially-variant kernels of each image, and (ii) a frequency-based kernel integration module implemented in a frequency domain-based Transformer backbone. \cref{fig:overall} shows the overview of the proposed method, details will be presented in the following.

\begin{figure*}[ht]
  \centering
  \includegraphics[width=1.0\linewidth]{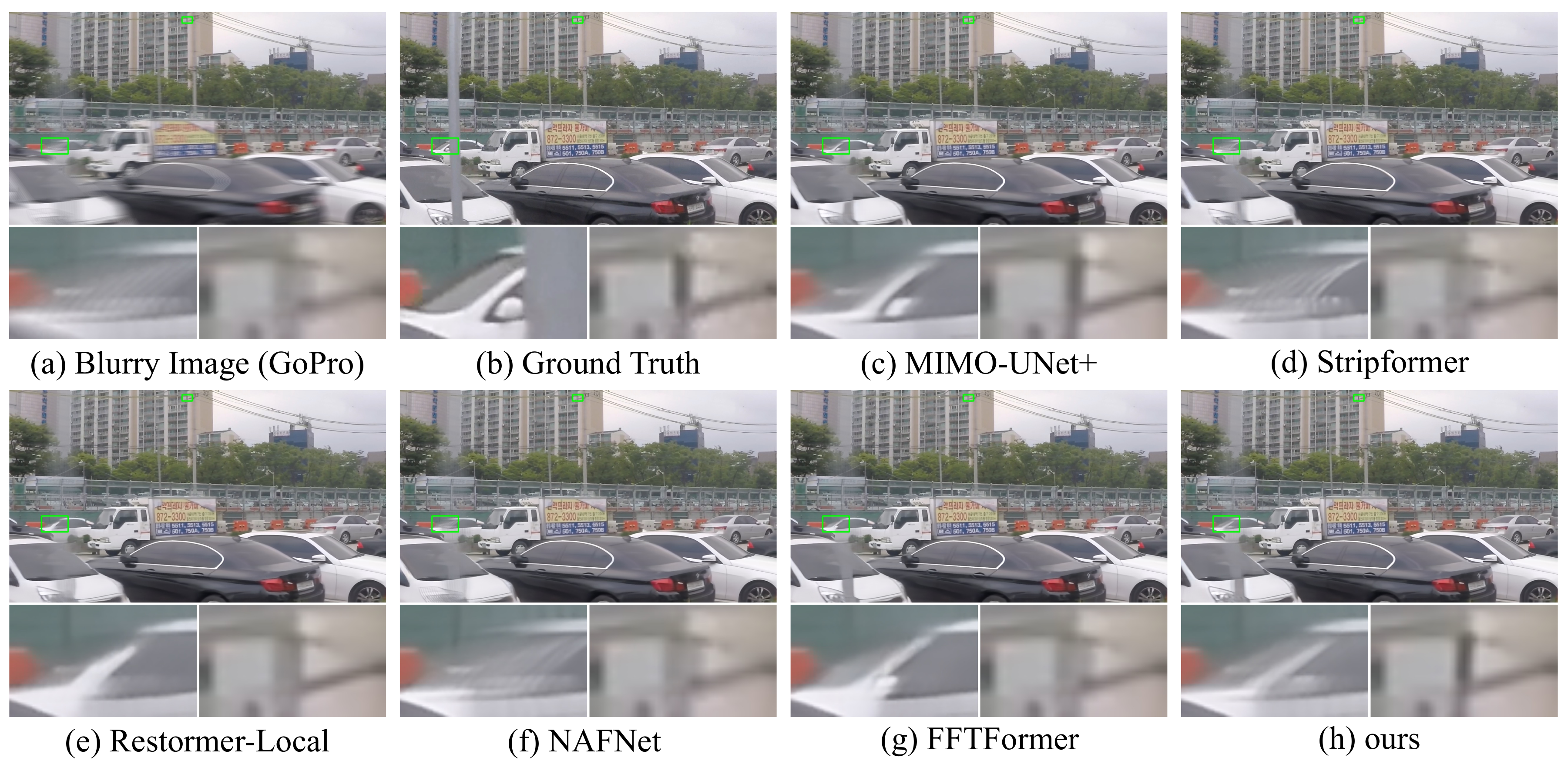}
  \hfill
\caption{Visual comparison on GoPro~\cite{nah2017deep} dataset. We compare our method with MIMO-UNet+~\cite{cho2021rethinking}, Stripformer~\cite{tsai2022stripformer}, Restormer-Local~\cite{zamir2022restormer}, NAFNet~\cite{chen2022simple}, and FFTFormer~\cite{kong2023efficient}. Models are trained only on the GoPro dataset. Our network generates more realistic images with clearer details.}
  \label{fig:gopro}
\end{figure*}

\subsection{Kernel Estimation Module}

Numerous methods have been employed in attempts to estimate the blur kernel from a blurry image. Most non-deep learning approaches typically employ optimization-based methods~\cite{levin2009understanding} to estimate the kernel, but they generally can only handle scenarios with spatially invariant kernels. The primary challenge faced by deep learning methods~\cite{bahat2017non} is the lack of ground truth. They typically use kernels obtained from optimization-based methods or synthesized kernels as supervision. The major issue with this approach is the difficulty of generalizing to real-world data. To address this issue, recent works like~\cite{kim2021koalanet} have proposed a self-supervised approach, estimating the kernel for real-world data by applying reblurring operations to sharp images. We adopt the method of  Kim et al.~\cite{kim2021koalanet} for our kernel estimation module.

Specifically, given the $n$-th pair of training data, a blurry image $\mathbf{y}_n \in \mathbb{R}^{H \times W \times 3}$ and its corresponding ground truth counterpart $\mathbf{x}_n \in \mathbb{R}^{H \times W \times 3}$, a convolutional network with the structure of U-Net, $f_{\mathbf{\theta}}$, is applied to predict the pixel-wise kernel, denoted as $\mathbf{B}_n$, using the blur image as input. Then the estimated kernel will be applied to reblur the sharp image, resulting in a pseudo blurry image. Subsequently, an $L_1$ Loss is adopted between the real blur image and the pseudo blurry image:

\begin{equation}
\label{eq:keloss}
\begin{split}
&\mathbf{B}_n = f_{\mathbf{\theta}}(\mathbf{y}_n),\\
&L_{KE} =  \frac{1}{N} \sum_{n=1}^{N} \| \mathbf{B}_n \otimes \mathbf{x}_n  - \mathbf{y}_n \|_1,
\end{split}
\end{equation}

where subscript $n$ represents the $n$-th pair of training data, $f_{\mathbf{\theta}}$ corresponds to the kernel estimation network with parameters $\mathbf{\theta}$, $\mathbf{B}_n$ is the predicted kernels by $f_{\mathbf{\theta}}$, $\otimes$ is the reblur operation, and $\|\cdot\|_1$ denotes the $L_1$ loss between two images.

\subsection{Frequency Integration Module}
With the given kernels, most conventional methods perform image deblurring through fast Fourier transform (FFT) based image deconvolution or minimizing the energy functions in closed forms. However, they often struggle to handle non-uniform scenarios and have shown inferior performance compared to deep learning methods in recent years. In order to implement the information of kernel prior into our backbone network, we propose \textbf{F}requency \textbf{I}ntegration \textbf{M}odule (FIM). FIM integrates the kernel prior into the Transformer-based structure in an implicit manner, fusing the feature and prior before each Transformer block, as shown in \cref{fig:overall} (c).


In each FIM block, the features of the image and kernel are fused in frequency domain via an operation similar to cross-attention mechanism, and an additional shortcut connection is applied to enable fast gradient propagation. Specifically, as shown in \cref{fig:overall} (b), FIM takes as input an image feature $\mathbf{X}_{in} \in \mathbb{R}^{\hat{C} \times \hat{H} \times \hat{W}}$, and a blur kernel feature $\mathbf{B}_{in} \in \mathbb{R}^{\hat{D} \times \hat{H} \times \hat{W}}$, where $\hat{H} \times \hat{W}$ is the spatial resolution of input features, ${\hat{C}}$ and ${\hat{D}}$ are channel numbers of the image features and kernel features, respectively. Given these feature inputs, we first project them into the same latent dimension via two consecutive spatial convolution and depth-wise convolution layers:

\begin{equation}
\label{eq:layer1}
\begin{split}
&\mathbf{Q} = {\rm DWConv_Q}({\rm Conv_Q}(\mathbf{X}_{in})),\\
&\mathbf{K} = {\rm DWConv_K}({\rm Conv_K}(\mathbf{B}_{in})),\\
&\mathbf{V} = {\rm DWConv_V}({\rm Conv_V}(\mathbf{B}_{in})),
\end{split}
\end{equation}

After obtaining $\mathbf{Q}(query)$, $\mathbf{K}(key)$ and $\mathbf{V}(value)$, we propose a new fusion method for the features, \textbf{F}requency \textbf{A}ttention (FA), which is an attention-like operation, implemented in the frequency domain. The operation of FA can be formulated as:
\begin{equation}
\label{eq:layer2}
\begin{split}
&\rm FA(\mathbf{Q}, \mathbf{K}, \mathbf{V})= \mathcal{L} (\mathcal{F}^{-1}(\mathcal{F}(\mathbf{Q}) \odot \mathcal{F}(\mathbf{K}))) \odot \mathbf{V},
\end{split}
\end{equation}
Where $\mathcal{F}$ denotes the Fourier transform, $\mathcal{F}^{-1}$ denotes the inverse Fourier transform, $\mathcal{L}$ represents layer normalization, and $\odot$ denotes element-wise multiplication of two tensors. Finally, we project the output of FA to match the feature dimension via a convolution layer, and add it to $\mathbf{X}_{in}$ in a residual way, obtaining the output feature $\mathbf{X}_{out}$:
\begin{equation}
\label{eq:layer3}
\mathbf{X}_{out} = {\rm Conv(FA(\mathbf{Q}, \mathbf{K}, \mathbf{V}))} + \mathbf{X}_{in},
\end{equation}

Moreover, since real-world motion blur consists of different sizes, it is insufficient to bring in kernel information at a single scale. To enable more efficient and precise injection of blur kernel prior information, we develop a multi-scale encoder-decoder integration strategy. 
Specifically, we generate multiple-scale blur kernel features $\mathbf{B}_{in1}, \mathbf{B}_{in2}, \mathbf{B}_{in3}$ by a downsampling operation, beginning with original kernel feature $\mathbf{B}_{in}$:
\begin{equation}
\label{eq:layer3}
\begin{split}
\mathbf{B}_{in1} &= \mathbf{B}_{in},\\
\mathbf{B}_{in2} &= {\rm Downsample}(\mathbf{B}_{in1}),\\
\mathbf{B}_{in3} &= {\rm Downsample}(\mathbf{B}_{in2}),\\
\end{split}
\end{equation}

Then, the downsampled features are integrated into layers with the corresponding dimension, both in the encoder and decoder. The effectiveness of our multi-scale encoder-decoder integration strategy is demonstrated in \cref{sec:ablation}.

\begin{table*}[ht!]
  \caption{Quantitative result comparisons on benchmark datasets GoPro~\cite{nah2017deep}, HIDE~\cite{shen2019human}, and RealBlur~\cite{rim2020real}. For all the evaluations, we use the model trained on the GoPro dataset for fair comparisons. Best and second best results are colored with {\color{red}red} and {\color{blue}blue}. Results show that our proposed method outperforms state-of-the-art methods on most image deblurring tasks.}
  \label{tab:all}
  \centering
  \setlength{\tabcolsep}{4mm}{
  \footnotesize
  \begin{tabular}{@{}l|cc|cc|cc|cc@{}}
    \toprule
    \multirow{2}{*}{Method}& \multicolumn{2}{c|}{GoPro}& \multicolumn{2}{c|}{HIDE} & \multicolumn{2}{c|}{RealBlur-R} & \multicolumn{2}{c}{RealBlur-J}\\
    & PSNR & SSIM & PSNR & SSIM& PSNR & SSIM& PSNR & SSIM \\
    \midrule\midrule
    DeblurGAN-v2~\cite{kupyn2019deblurgan}& 29.55 & 0.934 &26.61 & 0.875& 35.26 & 0.944 & 28.70 & 0.866\\
    SRN~\cite{tao2018scale} & 30.26 & 0.934& 28.36 & 0.915& 35.66 &0.947&28.56& 0.867\\
    DMPHN~\cite{zhang2019deep} & 31.20 & 0.945 & 29.09 & 0.924& 35.70 & 0.948&28.42& 0.860 \\
    SAPHN~\cite{suin2020spatially} & 31.85 &0.948 & 29.98 & 0.930 & - & -& - & -\\
    MIMO-UNet+~\cite{cho2021rethinking} & 32.45 &0.957 & 29.99 & 0.930& 35.54 & 0.947& 27.63& 0.837 \\
    MPRNet~\cite{zamir2021multi} & 32.66 &0.959& 30.96 & 0.939& 35.99 & 0.952&28.70 &0.873 \\
    DeepRFT+~\cite{mao2021deep} &33.23 &0.963&31.42 & 0.944&35.86 &0.950&28.97& {\color{blue}0.884} \\
    Restormer~\cite{zamir2022restormer} & 32.92 & 0.961 & 31.22 & 0.942& 36.19 & 0.957 & 28.96&0.879 \\
    HINet-Local~\cite{chu2022improving} &33.08 &0.962 & 30.66 & 0.936 & - &-&-&-\\
    Stripformer~\cite{tsai2022stripformer} & 33.08 &0.962 & 31.03 & 0.940& 36.07 & 0.952&28.82 & 0.876 \\
    MSDI-Net~\cite{li2022learning}& 33.28 &0.964& 31.02 & 0.940& 35.88& 0.952&28.59 &0.869\\
    MPRNet-Local~\cite{chu2022improving} & 33.31 &0.964 & 31.19 & 0.942 &-&-&-&-\\
    HI-Diff~\cite{chen2023hierarchical} &33.33 &0.964&31.46 & 0.945&{\color{blue}36.28} &{\color{red}0.958} & {\color{red}29.15} & {\color{red}0.890}\\
    Restormer-Local~\cite{zamir2022restormer} & 33.57 &0.966 & 31.49 & 0.945& - & - & -&- \\
    NAFNet~\cite{chen2022simple} & 33.71 &0.967&31.32 & 0.943&35.50& 0.953&28.32& 0.857\\
    GRL-B~\cite{li2023efficient} &33.93  &0.968&{\color{blue}31.65} & {\color{blue}0.947}&-&-&-&-\\
    FFTFormer~\cite{kong2023efficient} &{\color{blue}34.16} &{\color{red}0.969}& 31.62 & 0.946& 35.46& 0.952& 28.04& 0.852\\
    \hline
    Ours &{\color{red}34.21} &{\color{red}0.969} &{\color{red}32.08} &{\color{red}0.952}&{\color{red}36.59} &{\color{red}0.958} & {\color{blue}29.01}& 0.881\\
    \bottomrule
  \end{tabular}}
\end{table*}

\subsection{Overall Framework}

Inspired by Kong et al.~\cite{kong2023efficient}, we utilize their proposed network as the backbone of our proposed method to better fuse our frequency domain features. The authors of~\cite{kong2023efficient} propose a Transformer-based network for image deblurring in their work. They replace the matrix multiplication in the attention operation with frequency domain element-wise multiplication and develop a frequency domain-based feed-forward network. Here we name each block in our backbone as \textbf{F}requency \textbf{T}ransformer \textbf{B}lock(FTB).

The overall framework of our proposed network is shown in \cref{fig:overall}. Given a blurry image, the initial step involves feeding the blurry image into the kernel estimation module, which yields an estimated blur kernel for each pixel. Subsequently, these pixel-wise kernels are integrated into the deblurring network with the assistance of our proposed Frequency Integration Module. In our network design, FIM modules are inserted before each encoding and decoding Transformer block. 

The training schedule of the entire network unfolds as three distinct stages:
\begin{enumerate}
\item[I.] Preliminary training of the kernel estimation module, as outlined in \cref{eq:keloss};

\item[II.] Connect the kernel estimation network with the backbone via Frequency Integration Module, freeze the kernel estimation network, and train the backbone along with FIM. For the loss function $L_{phase2}$, we employ the same setting as in \cite{kong2023efficient}, shown in \cref{eq:p2loss}.
\begin{equation}
\label{eq:p2loss}
\begin{split}
&L_{cont} = \sum_{n=1}^{N} \frac{1}{N}\| \widehat{\mathbf{x}}_n - \mathbf{x}_n\|_1,\\
&L_{freq} = \sum_{n=1}^{N} \frac{1}{N}\|  \mathcal{F}(\widehat{\mathbf{x}}_n) -  \mathcal{F}(\mathbf{x}_n)\|_1,\\
&L_{phase2} = L_{cont} + \lambda L_{freq}.
\end{split}
\end{equation}

For the $n$-th pair of training data, $\widehat{\mathbf{x}}_n$ is the predicted sharp image by our network, and $\mathbf{x}_n$ is the sharp image ground truth. $\mathcal{F}$ stands for the Fourier transform. $L_{cont}$ calculates the $L_1$ loss between the deblurred image and ground truth, and $L_{freq}$ calculates their $L_1$ distance in the frequency domain. The total loss $L_{phase2}$ is the weighted sum of the above two losses;

\item[III.] Unfreeze the kernel estimation section and fine-tune the entire network, with an additional reblurring loss as supervision of the kernel estimation stage. The loss function $L_{phase3}$ can be expressed as \cref{eq:p3loss},
\begin{equation}
\label{eq:p3loss}
\begin{split}
&\mathbf{B}_n = f_{\mathbf{\theta}}(\mathbf{y}_n),\\
&L_{KE} =  \frac{1}{N} \sum_{n=1}^{N} \| \mathbf{B}_n \otimes \widehat{\mathbf{x}}_n  - \mathbf{y}_n \|_1,\\
&L_{phase3} = L_{cont} + \lambda_1 L_{freq} +  \lambda_2 L_{KE}.
\end{split}
\end{equation}

where $\mathbf{y}_n$ represents the blurry image in the $n$-th pair of training data, $f_{\mathbf{\theta}}$ is the kernel estimation network with parameters $\mathbf{\theta}$, $\mathbf{B}_n$ is the predicted kernel by $f_{\mathbf{\theta}}$, $\otimes$ is the reblur operation. $L_{KE}$ calculates the $L_1$ loss between the real blurry image and the reblurred image. $L_{cont}$ and $L_{freq}$ are the same as in stage 2, and the total loss $L_{phase3}$ is the weighted sum of the above three losses.
\end{enumerate}



\begin{figure*}[ht]
  \centering
  \includegraphics[width=1.0\linewidth]{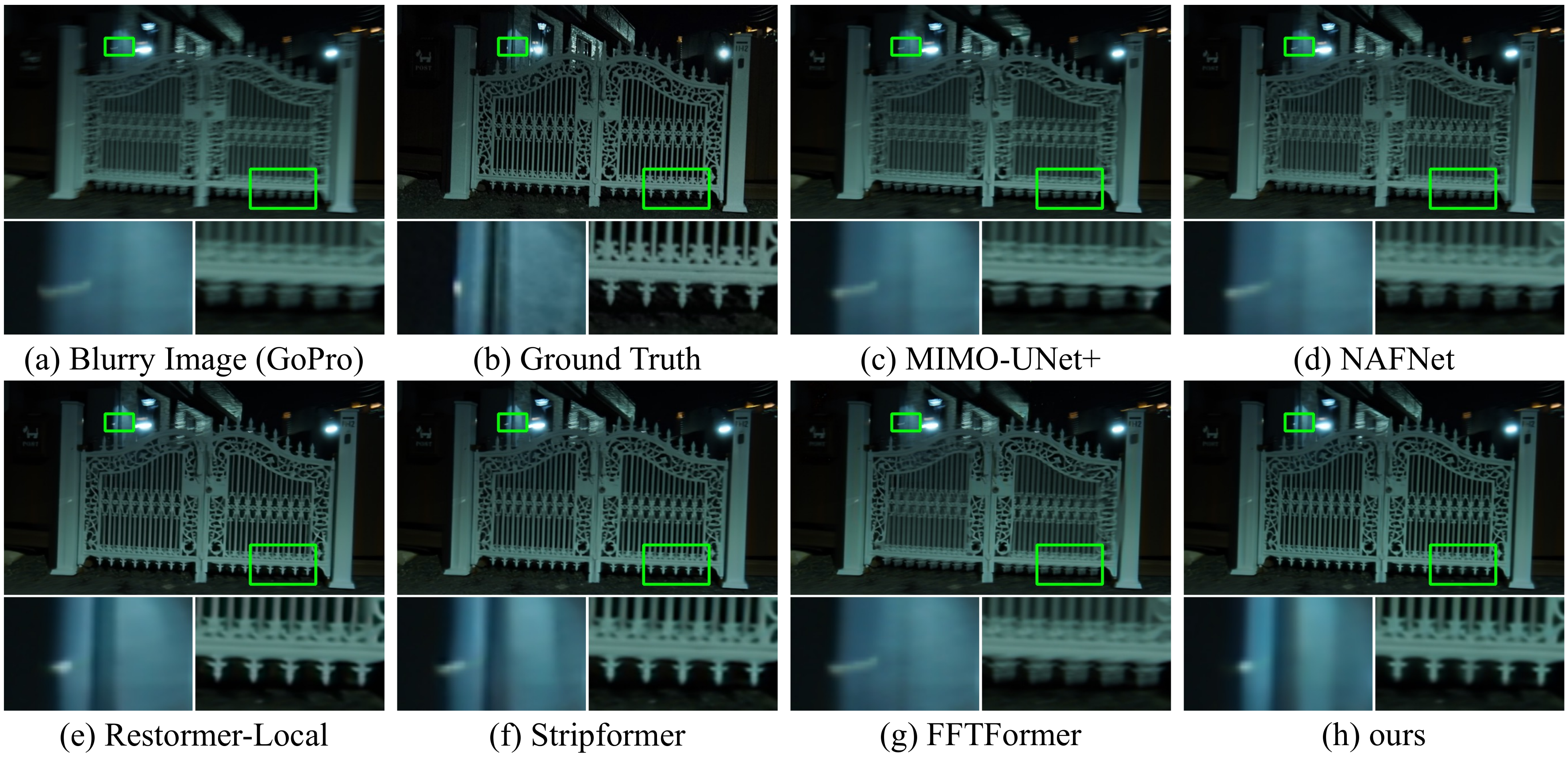}
  \hfill
  \caption{Visual comparison on RealBlur~\cite{rim2020real} dataset. We compare our method with MIMO-UNet+~\cite{cho2021rethinking}, NAFNet~\cite{chen2022simple}, Restormer-Local~\cite{zamir2022restormer}, Stripformer~\cite{tsai2022stripformer}, and FFTFormer~\cite{kong2023efficient}. Models are trained only on the GoPro dataset. Our network generates more realistic images with clearer details.}
  \label{fig:realblur1}
\end{figure*}
\section{Experiment}
\label{sec:experiment}
\subsection{Datasets and Implementation Details}

In alignment with the methods employed by prior blind image deblurring approaches~\cite{zamir2022restormer, kong2023efficient}, we train our proposed network using the extensive GoPro dataset~\cite{nah2017deep}, which consists of 2,103 pairs of blurry and sharp images for training. Subsequently, we conduct the evaluation of our method on both synthetic datasets (GoPro and HIDE~\cite{ shen2019human}) and a real-world dataset (RealBlur~\cite{ rim2020real}). GoPro dataset includes 1,111 test images, HIDE dataset provides 2,025 images for testing, and RealBlur consists of two datasets, namely RealBlur-R and RealBlur-J, both containing 980 blurry-sharp image pairs for testing. Following previous works~\cite{zamir2022restormer, kong2023efficient}, we adopt two common metrics, PSNR and SSIM~\cite{wang2004image}, for evaluation.

For training stage I, we train the kernel estimation module for 400,000 iterations with batch size of 16. At training stage II, we follow the training settings of \cite{kong2023efficient}. For the fine-tune training of stage III, Parameters $\lambda_1$ and $\lambda_2$ of $L_{phase3}$ in \cref{eq:p3loss} are both set to 0.1. We set the total iteration number to 200,000, all the other training parameters are the same as stage II.

Our proposed method is implemented by PyTorch. All experiments are conducted on 8 NVIDIA V100 GPUs.

\begin{figure*}[ht]

  \centering
  
  \includegraphics[width=1.0\linewidth]{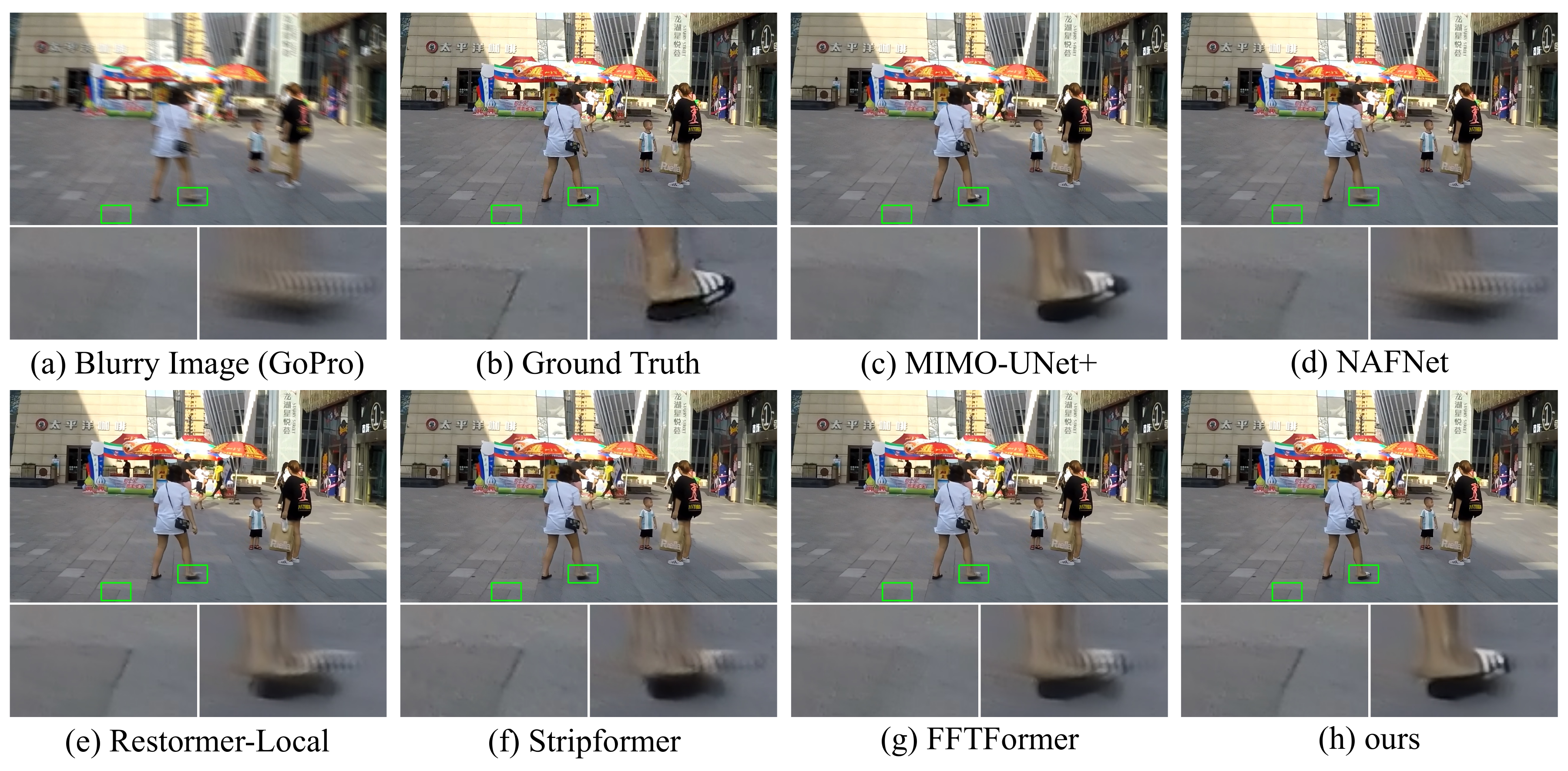}
  \hfill
\caption{Visual comparison on HIDE~\cite{shen2019human} dataset. We compare our method with MIMO-UNet+~\cite{cho2021rethinking}, NAFNet~\cite{chen2022simple}, Restormer-Local~\cite{zamir2022restormer}, Stripformer~\cite{tsai2022stripformer}, and FFTFormer~\cite{kong2023efficient}. Models are trained only on the GoPro dataset. Our network generates more realistic images with clearer details.}
  \label{fig:hide3}
\end{figure*}

\subsection{Comparisons with the state of the arts}

\subsubsection{Qualitative Comparison}

In this section, we compare our method with current state-of-the-art methods and use PSNR and SSIM metrics to analyze the quality of restored images. To ensure fair and unbiased comparisons, we retrain some of the previously proposed methods, following the same settings as the conventional training process on the corresponding dataset.

\textbf{I. Evaluations on GoPro.} We conduct training of our proposed network on the most commonly used GoPro~\cite{nah2017deep} dataset. Quantitative evaluation results on GoPro dataset are shown in \cref{tab:all}. Results show that our method generates the best results in terms of PSNR and SSIM. Note that we retrained the network proposed by Kong et al.~\cite{kong2023efficient} following the settings in the paper, but failed to obtain the stated result (0.05dB lower in PSNR). Therefore, we use our retrained results instead of the stated results for fair comparisons.


\textbf{II. Evaluations on HIDE.} We implement our model trained on GoPro dataset for evaluation on HIDE~\cite{ shen2019human} dataset. Quantitative evaluation results are shown in \cref{tab:all}. Our proposed method outperforms state-of-the-art methods with an advantage of 0.43dB in PSNR. This indicates that our blur kernel prior is able to enhance the model's generalization abilities with out-of-domain data. 


\textbf{III. Evaluations on RealBlur.} We implement our model trained on GoPro dataset for evaluation on RealBlur~\cite{ rim2020real} dataset. Quantitative evaluation results on RealBlur-R and RealBlur-J dataset are shown in \cref{tab:all}. The results show that our method outperforms state-of-the-art methods on RealBlur-R, while still maintaining considerable performance on RealBlur-J. This further demonstrates the superior generalization performance of our proposed method.

\textbf{IV. Computational Costs.}
We evaluate our methods by computational costs, specifically, in terms of model parameters, GFLOPS, and average runtime. Results are shown in \cref{tab:compute}. In comparison with other methods, our method consumes approximately equivalent time, while achieving better results, especially on out-of-domain data.
\begin{table}[h!]
  \caption{Quantitative evaluations on computational costs. The average runtime of all methods is tested on images with the size of $256 \times 256$ pixels.}
  \label{tab:compute}
  \centering
  \setlength{\tabcolsep}{3.0mm}{
  \footnotesize
  \begin{tabular}{@{}lccc@{}}
    \toprule
    Method & Parameters(M) & GFLOPs & Avg.runtime(s)\\
    \midrule
    MIMO-UNet+~\cite{cho2021rethinking} & 16.1 & 1235.3 & 0.02\\
    DeepRFT+~\cite{mao2021deep} & 23.0 & 187.0 & 0.09\\
    MPRNet-local~\cite{chu2022improving} & 20.1 & 778.2 & 0.11\\
    NAFNet~\cite{chen2022simple} & 67.8 & 65.0 & 0.04\\
    FFTFormer~\cite{kong2023efficient} & 16.6 & 131.8 & 0.13\\
    \hline
    Ours & 24.4 & 250.2 & 0.14\\
    \bottomrule
  \end{tabular}}
\end{table}

\subsubsection{Visual Comparison}

To further demonstrate our network's ability on out-of-domain data, we show visual comparisons on GoPro~\cite{nah2017deep}, RealBlur~\cite{rim2020real} and HIDE~\cite{shen2019human} in \cref{fig:gopro}, \cref{fig:realblur1} and \cref{fig:hide3}. The results show that our method can both restore highly degraded details and generate the most realistic visualization results.

\begin{table*}[h!]
  \caption{The ablation study of image deblurring results using different structures and implementation methods. All options are trained on GoPro~\cite{nah2017deep} dataset for the same iterations using the same setting. We test them on GoPro, HIDE~\cite{shen2019human}, RealBlur-R~\cite{rim2020real}, and RealBlur-J~\cite{rim2020real} datasets. \emph{Enc.} stands for Encoder, \emph{Dec.} stands for Decoder, \emph{Res.} stands for Residual connection. The first row denotes end-to-end training without introducing kernel prior, and the last row denotes the settings in our method.}
  \label{tab:ablation}
  \centering
  \setlength{\tabcolsep}{3.5mm}{
  \footnotesize
  \begin{tabular}{@{}cccc|c|c|c|c@{}}
    \toprule
    \multirow{2}{*}{Enc.}&\multirow{2}{*}{Dec.}&\multirow{2}{*}{Res.}&\multirow{2}{*}{Multiscale}& \multicolumn{4}{c}{PSNR / SSIM}\\
    \cline{5-8}
    &&&& GoPro& HIDE & RealBlur-R & RealBlur-J\\
    \midrule\midrule
    \XSolidBrush& \XSolidBrush & \XSolidBrush & \XSolidBrush& 32.45 / 0.954 & 30.48 / 0.934 & 34.49 / 0.939 & 27.62 / 0.855\\
    \CheckmarkBold & \XSolidBrush & \CheckmarkBold & \XSolidBrush & 32.97 / 0.960 & 31.12 / 0.942 & 35.37 / 0.952 & 27.95 / 0.861\\
    \XSolidBrush & \CheckmarkBold &  \CheckmarkBold &\XSolidBrush & 30.73 / 0.939 & 29.14 / 0.915 & 34.77 / 0.939 & 27.34 / 0.833\\
    \CheckmarkBold & \CheckmarkBold & \CheckmarkBold & \XSolidBrush& 33.03 / 0.961 & 31.12 / 0.942 & 35.15 / 0.951 & 28.05 / 0.862\\
     \CheckmarkBold & \CheckmarkBold &  \XSolidBrush& \CheckmarkBold & 30.37 / 0.936 & 29.09 / 0.915 & 34.22 / 0.935 & 27.02 / 0.829 \\
    \hline
    \CheckmarkBold & \CheckmarkBold & \CheckmarkBold &\CheckmarkBold & {\color{red}33.19} / {\color{red}0.964} & {\color{red}31.38} / {\color{red}0.944} & {\color{red}35.52} / {\color{red}0.955} & {\color{red}28.35} / {\color{red}0.871}\\
    \bottomrule
  \end{tabular}}
\end{table*}

\subsection{Ablation Studies}
\label{sec:ablation}
We conduct ablation studies to analyze our proposed methods in detail. The ablation studies include (i) experiments on different structures of Frequency Integration Module, for example, whether to use residual connection in FIM; (ii) different implementation strategies in the backbone network, for example, the position in the network where kernel information is introduced; and (iii) discussion on the non-blind version of our proposed method, where kernel ground truth is given. We perform these experiments on both in-domain and out-of-domain datasets.

\textbf{Effectiveness of encoder-decoder kernel implementation.} According to Kong et al.~\cite{kong2023efficient}, the backbone network adopts an asymmetric structure because the shallow features extracted by the encoder module usually contain blur effects that affect estimation. In our proposed overall framework, we integrate the kernel prior information through FIM, and insert the blocks in both the encoder and the decoder of the backbone. One may wonder if it’s more appropriate to implement FIM in a similar, asymmetric way. To address this assumption, we conduct ablation studies on the positions in which FIM modules are implemented. We try (i) only introducing prior in the encoder; (ii) only introducing prior in the decoder, and (iii) introducing prior in both the encoder and decoder. Experimental results are shown in row~2-4 of \cref{tab:ablation}. The results prove that introducing kernel prior in both branches brings the best effects in deblurring. The results also show that using kernel prior in the encoder proves more useful than in the decoder, indicating that degradation removal inside the network is mostly conducted in the encoder.

\textbf{Effectiveness of multiscale implementation.} In this part we explore the effectiveness of the multiscale implementation strategy. Experimental results are shown in row~4 and row~6 of \cref{tab:ablation}. For rows without `multiscale' ticked, we implement the kernel prior in a single-scale way, where we insert Frequency Integration Module (FIM) block only before the block with the largest scale (the first and the last). Results show that multi-scale implementation helps our network with deblurring, both on in-domain datasets, and on out-of-domain datasets.

\textbf{Effectiveness of residual connection.} We adopt the structure of residual connection in our proposed FIM block, demonstrated in \cref{fig:overall} (b). To prove the effect of residual connection in our block, we remove the residual connection in the FIM block and train the entire network with the same parameter settings. As shown in \cref{tab:ablation}, the results in row~5 and row~6 give evidence that residual connection plays an important role in the function of FIM. Without a residual connection, FIM module could have a negative influence on the deblurring ability of the network.

\begin{table}[!h]
   \caption{Quantitative comparisons between blind and non-blind methods. \iffalse Kernel ground truth is given under non-blind settings. \fi\emph{Baseline} denotes end-to-end training without introducing kernel prior, \emph{Ours} is our original method, and \emph{Ours (non-blind)} is the non-blind version of our method, which means that kernel ground truth is given. All methods are trained on GoPro~\cite{nah2017deep} dataset and tested on our four synthetic datasets: \emph{GoPro (Syn.)}, \emph{HIDE (Syn.)}, \emph{RB-R (Syn.)}, and \emph{RB-J (Syn.)}, respectively based on GoPro, HIDE~\cite{shen2019human}, RealBlur-R~\cite{rim2020real} and RealBlur-J~\cite{rim2020real}.}
  \label{tab:blind}
  \centering
  \setlength{\tabcolsep}{0.5mm}{
  \footnotesize
  \begin{tabular}{@{}l|c|c|c|c@{}}
    \toprule
    \multirow{2}{*}{Method} & \multicolumn{4}{c}{PSNR/SSIM}\\
    \cline{2-5}
    & GoPro (Syn.)& HIDE (Syn.) & RB-R (Syn.) & RB-J (Syn.)\\
    \midrule\midrule

    
    Baseline &  28.48 / 0.918 & 27.35 / 0.884 & 37.95 / 0.968 &29.98 / 0.916 \\
    Ours & 28.59 / 0.919 & 27.73 / 0.893 & 38.01 / 0.970 &30.44 / 0.919 \\
    Ours(non-blind) & \textbf{28.80} / \textbf{0.922} & \textbf{27.88} / \textbf{0.896} & \textbf{38.08} / \textbf{0.970} &\textbf{30.51} / \textbf{0.920} \\

    \bottomrule
  \end{tabular}}
\end{table}
\textbf{Discussion on the non-blind version of our method.} Our proposed module FIM helps with deblurring when given estimated blur kernels. To further verify the effectiveness of the module, we replace the predicted kernels of the kernel estimation stage with ground truth kernels. We perform the ablation study on synthetic datasets with ground truth kernels. The synthetic datasets are respectively based on GoPro~\cite{nah2017deep}, HIDE~\cite{shen2019human}, RealBlur-R~\cite{rim2020real} and RealBlur-J~\cite{rim2020real}. We first extract the kernels of each blurry image in the original dataset, and add these kernels using reblur operation to their corresponding sharp images, obtaining the synthetic dataset. The tested results are shown in \cref{tab:blind}. From the results we can see that an accurate blur kernel can further help our network with deblurring, indicating that our method can benefit from a more accurate kernel estimator, enabling further improvements.



\section{Conclusion}
\label{sec:conclusion}

\quad In this paper, we introduce an innovative approach that leverages explicit blur kernel priors to achieve effective image deblurring. To address the issue of out-of-domain data distribution, we propose a way of implementing explicit blur kernel prior to a Transformer-based network. Extensive experimental results show that the incorporation of kernel information enhances the network's deblurring capability. Notably, our approach surpasses the current state-of-the-art methods on both in-domain and out-of-domain deblurring tasks, further validating the efficacy and robust generalization ability of our proposed method. Limitations in our work like the computation cost are left for further studies in the future. 

{
    \small
    \bibliographystyle{ieeenat_fullname}
    \bibliography{main}
}


\end{document}